%% file: main.tex
\pdfoutput=1

\documentclass[11pt]{article}

\usepackage{acl2023} 

\usepackage{times}
\usepackage{latexsym}

\usepackage[T1]{fontenc}

\usepackage[utf8]{inputenc}

\usepackage{microtype}

\usepackage{inconsolata}

\usepackage{graphicx}
\usepackage{float}

\title{\model{}: AMR-Aware Prefix for Generation-Based \\ Event Argument Extraction Model}

\author{I-Hung Hsu\thanks{\; The authors contribute equally.}$^{\;\;1}$ \ \ \ Zhiyu Xie\footnotemark[1]$^{\;\;3}$ \ \ \ Kuan-Hao Huang$^{2}$ \\
{\bf Premkumar Natarajan$^{1}$ \ \ \ Nanyun Peng$^{2}$}\\
$^{1}$Information Science Institute, University of Southern California \\
$^{2}$Computer Science Department, University of California, Los Angeles \\ 
$^{3}$Computer Science Department, Tsinghua University \\ 
\texttt{\{ihunghsu, pnataraj\}@isi.edu} ~~
\texttt{\{khhuang, violetpeng\}@cs.ucla.edu} \\
\texttt{xiezy19@mails.tsinghua.edu.cn} \\}
\input{macro}

\begin{document}
\maketitle

\input{00-abstract}
\input{01-intro}
\input{03-method}
\input{04-experiments}
\input{05-studies}
\input{02-related}
\input{06-conclusion}

\bibliography{anthology,custom}
\bibliographystyle{acl_natbib}

\input{99-appendix}

\end{document}

%% file: macro.tex
\usepackage[flushleft]{threeparttable}
\usepackage{todonotes}
\usepackage{enumitem}
\usepackage{booktabs}
\usepackage{subcaption}
\usepackage[nointegrals]{wasysym}
\usepackage{tikz}
\usepackage{tabularx}
\usepackage{tabulary}
\usepackage{listings}
\usepackage{color}
\usepackage{multirow}
\usepackage{colortbl}
\usepackage{framed}
\usepackage{pifont}
\usepackage{dsfont}
\usepackage{xcolor}
\usepackage{amsmath,amsthm}
\usepackage{amssymb}
\usepackage{graphicx}
\usepackage{textcomp}
\usepackage{tcolorbox}
\usepackage{soul}
\usepackage{lipsum}
\usepackage{makecell}
\usepackage{amsfonts}
\usepackage{bbm}
\setuldepth{Berlin}

\definecolor{DarkRed}{RGB}{130,25,0}

\newcommand{\ignore}[1]{}

\definecolor{mypurple}{RGB}{121,55,196}
\definecolor{myblue}{RGB}{60,177,245}
\definecolor{myorange}{RGB}{243,206,84}
\definecolor{mygreen}{RGB}{41,222,35}
\definecolor{light-gray}{gray}{0.9}

\newcommand{\Skip}[1]{{}}
\newcommand{\model}[1]{\textsc{Ampere}}
\newcommand{\modelbart}[1]{\textsc{Ampere} (\texttt{AMRBART)}}
\newcommand{\modelroberta}[1]{\textsc{Ampere} (\texttt{RoBERTa)}}
\newcommand{\modeldegree}[1]{\textsc{DEGREE}}
\newcommand{\modelwocopy}[1]{\textsc{Ampere}_{-copy}}
\newcommand{\degree}[1]{\textsc{Degree}}

\newcommand{\AMRPrefix}[1]{AMR-aware Prefix}
\newcommand{\anycopy}[1]{any copy}
\newcommand{\purecopy}[1]{pure copy}
\newcommand{\noregcopy}[1]{ordinary copy mechanism}

\usepackage{cleveref}
\crefformat{section}{Section~#2#1#3}
\crefformat{subsection}{Section~#2#1#3}
\crefformat{subsubsection}{Section~#2#1#3}
\crefrangeformat{section}{\S\S#3#1#4 to~#5#2#6}
\crefmultiformat{section}{\S\S#2#1#3}{ and~#2#1#3}{, #2#1#3}{ and~#2#1#3}
\crefmultiformat{subsection}{\S\S#2#1#3}{ and~#2#1#3}{, #2#1#3}{ and~#2#1#3}

\Crefformat{figure}{#2Figure~#1#3}
\Crefmultiformat{figure}{Figures~#2#1#3}{ and~#2#1#3}{, #2#1#3}{ and~#2#1#3}
\Crefformat{table}{#2Table~#1#3}
\Crefmultiformat{table}{Tables~#2#1#3}{ and~#2#1#3}{, #2#1#3}{ and~#2#1#3}
\Crefformat{appendix}{Appendix~\S#2#1#3}
\crefmultiformat{appendix}{Appendix~\S#2#1#3}{ and~#2#1#3}{, #2#1#3}{ and~#2#1#3}
\crefformat{algorithm}{Algorithm~#2#1#3}
\Crefformat{equation}{Equation~#2#1#3}

%% file: 00-abstract.tex
\begin{abstract}
Event argument extraction (EAE) identifies event arguments and their specific roles for a given event.
Recent advancement in generation-based EAE models has shown great performance and generalizability over classification-based models. 
However, existing generation-based EAE models mostly focus on problem re-formulation and prompt design, without incorporating additional information that has been shown to be effective for classification-based models, such as the abstract meaning representation (AMR) of the input passages. 
Incorporating such information into generation-based models is challenging due to the heterogeneous nature of the natural language form prevalently used in generation-based models and the structured form of AMRs.
In this work, we study strategies to incorporate AMR into generation-based EAE models. We propose \model{}, which generates AMR-aware prefixes for every layer of the generation model.
Thus, the prefix introduces AMR information to the generation-based EAE model and then improves the generation.
We also introduce an adjusted copy mechanism to \model{} to help overcome potential noises brought by the AMR graph.
Comprehensive experiments and analyses on ACE2005 and ERE datasets show that \model{} can get $4\% - 10\%$ absolute F1 score improvements with reduced training data and it is in general powerful across different training sizes.
\looseness=-1

\end{abstract}

%% file: 01-intro.tex
\section{Introduction}
\label{sec:intro}

\begin{figure*}[t!]
\centering 
\includegraphics[width=0.99\textwidth]{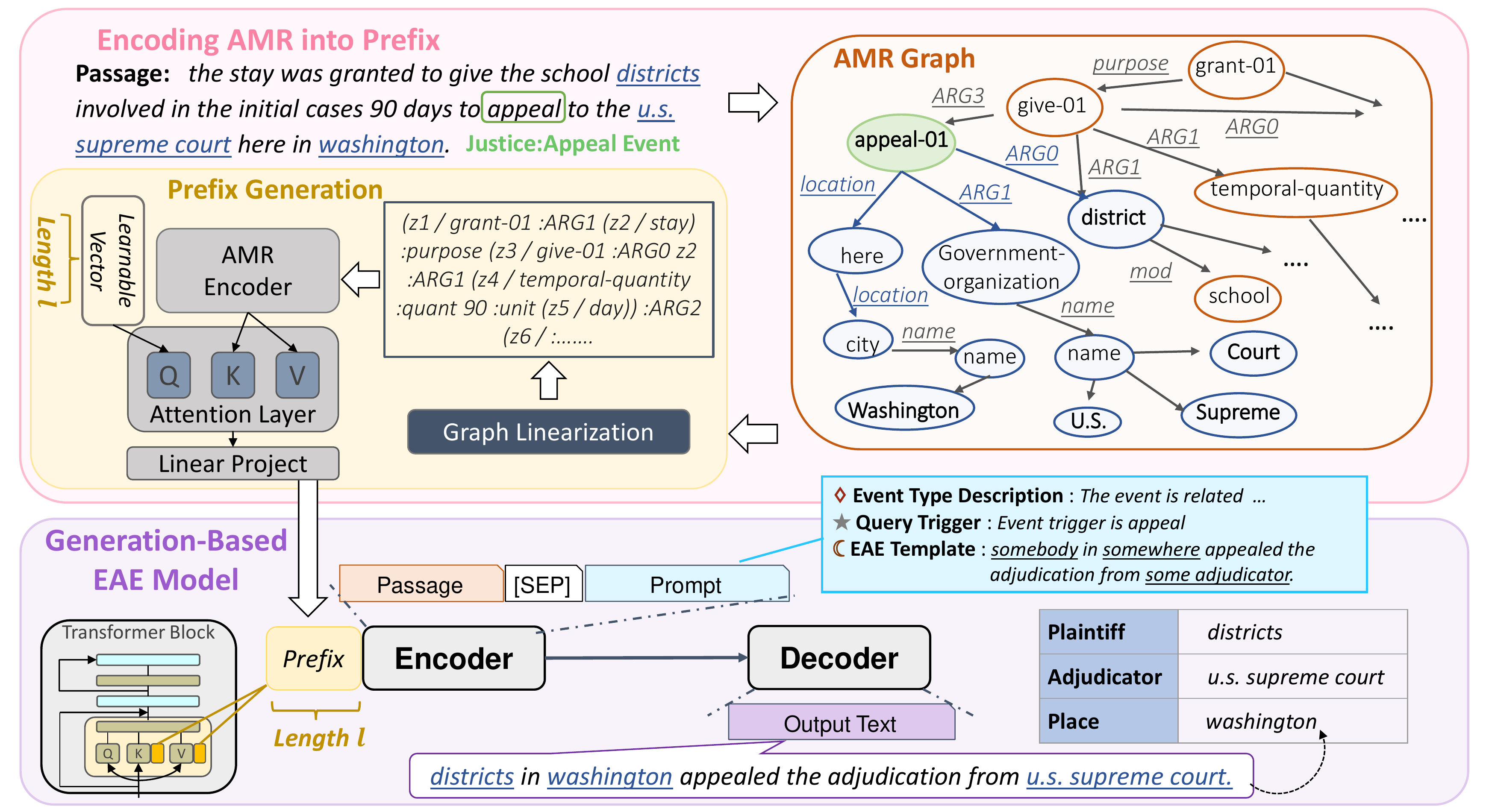} 
\caption{An overview of \model{} using an example from the ACE 2005 dataset. Given a passage and an event trigger, we first use an AMR parser to obtain the AMR graph of the input passage. The linearized AMR graph sequence will be encoded into a $l$-length prefix by an AMR encoder and an attention layer. Our generation-based EAE model equipped with the AMR-aware prefix then summarizes the event mentioned in the passage into a natural sentence that follows a pre-defined template in the prompt. The final arguments and the corresponding roles can be extracted from the generated sentence.} 
\label{fig:model} 
\end{figure*}

Event argument extraction (EAE) aims to recognize event arguments and their roles in an event. For example, in \Cref{fig:model}, EAE models need to extract \textit{districts}, \textit{u.s. supreme court}, and \textit{washington} and the corresponding roles --- \textit{Plaintiff}, \textit{Adjudicator}, and \textit{Place} for the \textit{Justice:Appeal} event  with trigger \textit{appeal}. EAE has long been a challenging task in NLP, especially when training data is limited~\cite{DBLP:conf/emnlp/WangWHLLLSZR19, ma-etal-2022-prompt}. It is an important task for various downstream applications~\cite{DBLP:conf/www/ZhangLPSL20, berant-etal-2014-modeling, hogenboom2016survey, DBLP:conf/naacl/WenLLPLLZLWZYDW21, DBLP:conf/naacl/0001HFJ22}.

Recently, generation-based EAE models~\cite{Hsu22degree, DBLP:conf/acl/0001LXHTL0LC20, DBLP:conf/naacl/LiJH21, DBLP:conf/iclr/PaoliniAKMAASXS21, DBLP:journals/corr/abs-2205-12505} are proposed and have shown great generalizability and competitive performance compared to traditional classification-based methods~\cite{DBLP:conf/acl/ChenXLZ015, DBLP:conf/emnlp/MaWABA20, DBLP:journals/corr/abs-2205-12585, DBLP:conf/aaai/FinckeAMB22}.
However, existing generation-based EAE models mostly focus on problem reformulation and prompt design without 
incorporating auxiliary syntactic and semantic information that is shown to be effective in  
classification-based methods~\cite{DBLP:conf/acl/HuangCFJVHS16, DBLP:journals/corr/abs-2205-12490, DBLP:conf/acl/DaganJVHCR18, DBLP:conf/aaai/AhmadPC21, DBLP:conf/emnlp/VeysehNN20}.

In this work, we explore how to incorporate auxiliary structured information into generation-based EAE models. We focus on abstract meaning representation (AMR)~\cite{DBLP:conf/acllaw/BanarescuBCGGHK13}, which extracts rich semantic information from the input sentence. 
As the \Cref{fig:model}'s example shows, AMR graph summarizes the semantic structure of the input passage, and many of its nodes and edges share strong similarities with the event structures. For example, the trigger word \textit{appeal} can be mapped to the node \textit{``appeal-01''}, and the subject who appeals can be found using edge \textit{``ARG0''}.
Hence, the AMR graph could provide important clues for models to figure out event arguments, resulting in performance improvements~\cite{DBLP:conf/naacl/ZhangJ21} and better generalizability~\cite{DBLP:conf/acl/DaganJVHCR18} for classification-based methods. However, it is unclear how to best integrate AMR into generation-based methods. 
The heterogeneous nature between the AMR graph and the natural language prompts\footnote{For example, event type description and target generation templates.} in the generation-based EAE models causes the difficulty of the model design.

To overcome the challenge, we 
propose \model{} (\textbf{AM}r-aware \textbf{P}refix for generation-based \textbf{E}vent a\textbf{R}gument \textbf{E}xtraction), which encodes AMR graph into prefix~\cite{prefixtuning} to regulate the generation-based EAE models. 
Specifically, an additional AMR encoder is used to encode the input AMR graph into dense vectors. Then, these vectors will be disassembled and distributed to every Transformer layer in generation-based EAE models as the prefix. These generated prefixes are transformed into additional key and value matrices to influence the attention calculation, hence, guiding the generation.


We also introduce an adjusted copy mechanism for \model{} to overcome potential noises brought by the AMR graph. Specifically, as we can observe in \Cref{fig:model}, AMR parsers will include additional normalization (turning \textit{washington} into \textit{Washington}) and word disambiguation (using \textit{appeal-01} rather than \textit{appeal}) to create AMR graphs. Such normalization could impact the generation to produce some words that are not in the original input, especially when the training data is limited. Hence, we apply a copy mechanism~\cite{DBLP:conf/acl/SeeLM17} and add an additional regularization loss term to encourage copying from the input passage. 

We conduct experiments on ACE 2005 \cite{doddington-etal-2004-automatic} and ERE \cite{DBLP:conf/aclevents/SongBSRMEWKRM15} datasets using different ratios of training data. Our results show that \model{} outperforms several prior EAE works in both datasets. Under low-resource settings that use only $5\%$ or $10\%$ of training data, we can get $4\% - 10\%$ absolute F1-scores of improvement, and our method is in general powerful across different training sizes and different datasets. We also present a comprehensive study of different ways to incorporate AMR information into a generation-based EAE model. We will show that \model{} is the best way among the various methods we explored. Our code can be found at \url{https://github.com/PlusLabNLP/AMPERE}.

%% file: 03-method.tex
\section{Method}
\label{sec:method}
\model{} uses \degree{}~\cite{Hsu22degree} as the base generation-based EAE model~\footnote{We use the EAE version of \degree{}~\cite{Hsu22degree}.} (\Cref{subsec:Degree}), and augments it with AMR-aware prefixes, as shown in \Cref{fig:model}.
To generate the AMR-aware prefixes, we first use a pre-trained AMR parser to obtain the AMR graph of the input sentence (\Cref{subsec:amr-parsing}).
Then, the graph is transformed into dense vectors through graph linearization and an AMR encoder. Then, these dense vectors will be disassembled and distributed to each layer of our base generation-based EAE model so the generation is guided by the AMR information (\Cref{subsec:amr-prefix}).
Finally, we introduce the training loss for \model{} and our adjusted copy mechanism that can help \model{} overcome additional noise brought from AMR graphs (\Cref{subsec:copy}).

\subsection{Generation-Based EAE Model}
\label{subsec:Degree}
Despite our AMR-aware prefix being agnostic to the used generation-based EAE model, we select \degree{}~\cite{Hsu22degree} as our base model because of its great generalizability and performance. Here, we provide a brief overview of the model.

Given a passage and an event trigger, \degree{} first prepares the \textit{prompt}, which includes an event type description (a sentence describing the trigger word), and an event-type-specific template, as shown in \Cref{fig:model}. Then, given the passage and the prompt, \degree{}  summarizes the event in the passage following the format of the EAE template, so that final predictions can be decoded easily by comparing the template and the output text. Take the case in \Cref{fig:model} as an example, by comparing \textit{``\ul{districts} in \ul{washington} appealed the adjudication from \ul{u.s. supreme court}.''} with the template \textit{``\ul{somebody} in \ul{somewhere} appealed the adjudication from \ul{some adjudicator}.''}, we can know that the ``districts'' is the argument of role \textit{``Plaintiff''}. This is because the corresponding placeholder \textit{``somebody''} of the role \textit{``Plaintiff''} has been replaced by \textit{``districts''} in the model's prediction.

\subsection{AMR Parsing}
\label{subsec:amr-parsing}
The first step of our method is to prepare the AMR graph of the input passage. We consider SPRING \cite{Bevilacqua21amrparse1}, a BART-based AMR parser trained on AMR 3.0 annotation,~\footnote{\url{https://catalog.ldc.upenn.edu/LDC2020T02}} to be our AMR parser. As illustrated by \Cref{fig:model}, the AMR parser encodes the input sentence into an AMR graph, which is a directed graph where each node represents a semantic concept (e.g., \textit{``give-01''}, \textit{``appeal-01''}) and each edge describe the categorical semantic relationship between two concepts (e.g., \textit{ARG0}, \textit{location})~\cite{DBLP:conf/acllaw/BanarescuBCGGHK13}.

\subsection{AMR-Aware Prefix Generation}
\label{subsec:amr-prefix}
Our next step is to embed the information into prefixes~\cite{prefixtuning} for our generation-based EAE model. To encode the AMR graph, we follow \citet{DBLP:conf/acl/KonstasIYCZ17} to adopt a depth-first-search algorithm to linearize the AMR graph into a sequence, as shown in the example in \Cref{fig:model}. Then, an AMR encoder is adapted to encode the representation of the sequence. One of the advantages of our method is the flexibility to use models with different characteristics to our generation-based EAE model to encode AMR.
Here, we consider two AMR encoders to form different versions of \model{}:
\begin{itemize} [topsep=3pt, itemsep=-2pt, leftmargin=13pt]
    \item \modelbart{}: We consider using the encoder part of the current state-of-the-art AMR-to-text model --- \texttt{AMRBART}~\cite{DBLP:conf/acl/00010022} that pre-trained on AMR 3.0 data.~\footnote{\url{https://github.com/goodbai-nlp/AMRBART}} The model is based on \texttt{BART-large} and its vocabulary is enlarged by adding all relations and semantic concepts in AMR as additional tokens. Employing the model as our AMR encoder enables \model{} to leverage knowledge from other tasks. 
    \item \modelroberta{}: \texttt{RoBERTa-large}~\cite{DBLP:journals/corr/abs-1907-11692} is also considered as our AMR encoder as pre-trained masked language models are typical choices to perform encoding tasks.
    In order to make \texttt{RoBERTa} better interpret the AMR sequence, we follow \citet{DBLP:conf/acl/00010022} to add all relations in AMR (e.g. \textit{ARG0}, \textit{ARG1}) as special tokens. However, since the model is not pre-trained on abundant AMR-to-text data, we do not include semantic concepts (e.g. concepts end with \textit{-01}) as extra tokens.~\footnote{If adding semantic concepts as extra tokens, \texttt{RoBERTa} will loss the ability to grasp its partial semantic meaning from its surface form, such as understanding that \textit{``appeal-01''} is related to \textit{``appeal''}.} 
\end{itemize}

After getting the representation of the linearized sequence, we then prepare $l$ learnable vectors as queries and an attention layer, where $l$ is a hyper-parameter that controls the length of the used prefixes. These queries will compute attention with the representations of the linearized AMR sequence, then, we will obtain a set of compressed dense vector $\mathbf{P}$. This $\mathbf{P}$ will be transformed into the prefixes~\cite{prefixtuning} that we will inject into our generation-based EAE model.

To be more specific, we first disassemble $\mathbf{P}$ into $L$ pieces, where $L$ is the number of layers in the base generation-based EAE model, i.e., $\mathbf{P}=\{P^1, P^2, ...P^L\}$. Then, in the $n$-th layer of the EAE model, the prefix is separated into two matrices, standing for the addition key and value matrices: $P^n = \{K^n, V^n\}$, where $K^n$ \& $V^n$ are the addition key and value matrices, and they can be further written as $K^n=\{\mathbf{k}^n_1, ..., \mathbf{k}^n_l\}$ and $V^n=\{\mathbf{v}^n_1, ..., \mathbf{v}^n_l\}$. $\mathbf{k}_*$ and $\mathbf{v}_*$ are vectors with the same hidden dimension in the Transformer layer. These additional key and value matrices will be concatenated with the original key and value matrices in the attention block. Therefore, when calculating dot-product attention, the query at each position will be influenced by these AMR-aware prefixes. The reason of generating layer-wise queries and keys is to exert stronger control. We generate layer-wise key-value pairs as each layer may embed different information. These keys influence the model's weighting of representations towards corresponding generated values. Empirical studies on layer-wise versus single-layer control can be found in \citet{DBLP:conf/acl/LiuJFTDY022}. 

It is worth noting that \citet{prefixtuning}'s prefix tuning technique uses a fixed set of prefixes disregarding the change of input sentence, \model{} will \textit{generate} a different set of prefixes when the input passage varies. And the variation reflects the different AMR graph's presentation.

We can inject prefixes into the encoder self-attention blocks, decoder cross-attention blocks, or decoder self-attention blocks in our generation-based EAE model. Based on our preliminary experiments, we observe that using prefix in encoder self-attention blocks and decoder cross-attention blocks works best in \model{}.

\subsection{Adjusted Copy Mechanism}
\label{subsec:copy}
We follow \degree{}'s setting to use \texttt{BART-large}~\cite{DBLP:conf/acl/LewisLGGMLSZ20} as the pre-trained generative model, and the training objective of our generation-based EAE model is to maximize the conditional probability of generating a ground-truth token given the previously generated ones and the input context in the encoder $x_1, x_2, ..x_m$:
\begin{equation}
\label{equation:normal-loss}
    \small Loss = -\log (\sum_{i} P(y_i|y_{<i}, x_1, ..., x_m)),
\end{equation}
where $y_i$ is the output of the decoder at step $i$.
In \degree{}'s setting, the probability of predicting an token $t$ fully relies on the generative model. 
Although this setting is more similar to how \texttt{BART-large} is pre-trained and thus better leverages the power of pre-training, the loose constraints on the final prediction could generate hallucinated texts~\cite{DBLP:journals/corr/abs-2202-03629} or outputs not following the template. Such an issue could be enlarged if less training data is used and more input noise is presented, such as when incorporating AMR graphs.

To enhance the control, one commonly-used technique is to apply copy mechanism~\cite{DBLP:conf/acl/SeeLM17} to generation-based event models~\cite{DBLP:conf/acl/HuangHNCP22, huang-etal-2021-document}.
, i.e.,
\begin{equation}
\label{equation:copy}
\small
\begin{split}
P(&y_i=t|y_{<i}, x_1, .., x_m) =\\
&w^i_{gen}P_{gen}(y_i=t|y_{<i}, x_1, ..., x_m)) + \\
 &(1-w^i_{gen})(\sum_{j=0}^{m}P^i_{copy}(j|y_{<i}, x_1, ..., x_m) \times \mathbbm{1}(x_j=t)),
\end{split}
\end{equation}
where $w^i_{gen}\in[0,1]$ is the probability to generate, computed by passing the last decoder hidden state to an additional network. 
$P^i_{copy}(j|\cdot)$ is the probability to copy input token $x_j$, and it's computed by using the cross-attention weights in the last decoder layer at time step $i$. When $w^i_{gen}=1$, it is the original model used by \degree{}, while if $w^i_{gen}=0$, this model will only generate tokens from the input.

Our core idea of the adjusted copy mechanism is to encourage the model to copy more, and this is achieved by introducing a regularization term on $w^i_{gen}$ to the loss function of \model{}:
\begin{equation}
\label{equation:copy-with-reg}
\small
\begin{split}
Loss_{\model{}} =& \\ 
-\log (\sum_{i} &P(y_i|y_{<i}, x_1, ..., x_m)) + \lambda\sum_{i}w^i_{gen},
\end{split}
\end{equation}
where $\lambda$ is a hyper-parameter.
Compared to fully relying on copy from input, our method still allows the generative model to freely generate tokens not presented in the input. 
Compared to ordinary copy mechanisms, the additional regularizer will guide the model to copy more.
Using this loss, we train the whole \model{} end-to-end.

\begin{table*}[t!]
\centering
\small
\resizebox{1.0\textwidth}{!}{
\setlength{\tabcolsep}{3.5pt}
\renewcommand{\arraystretch}{1.05}
\begin{tabular}{lccccccc|cccccc}
    \toprule
    \multirow{2}{*}{Model} & \multirow{2}{*}{Type} & \multicolumn{6}{c}{Development Set} & \multicolumn{6}{c}{Test Set} \\
    \cmidrule{3-14}
    & & 5\% & 10\% & 20\% & 30\% & 50\% & 100\% & 5\% & 10\% & 20\% & 30\% & 50\% & 100\%  \\
    \midrule
    \multicolumn{14}{c}{\textbf{Argument Classification F1-Score (\%) in ACE05-E}} \\
    \midrule
    DyGIE++~\cite{Wadden19dygiepp}& Cls
    & 34.6 & 48.5 & 52.5 & 57.5 & 57.9 & 60.0
    & 29.3 & 42.4 & 49.5 & 53.2 & 54.5 & 57.4 \\
    OneIE~\cite{Lin20oneie}& Cls
    & 38.6 & 56.0 & 63.2 & 67.6 & 70.4 & 71.8
    & 34.6 & 50.0 & 59.6 & 63.0 & 68.4 & 70.6 \\
    Query and Extract~\cite{wang-etal-2022-query}& Cls
    & 10.5 & 27.7 & 37.6 & 50.0 & 54.6 & 61.7
    & 11.0 & 20.9 & 34.3 & 44.3 & 49.6 & 59.1 \\
    AMR-IE~\cite{DBLP:conf/naacl/ZhangJ21}& Cls
    & 40.0 & 56.3 & 61.3 & 67.4 & 70.6 & 73.1
    & 36.8 & 48.5 & 58.3 & 62.6 & 66.1 & 70.3 \\
    PAIE~\cite{ma-etal-2022-prompt}& Gen
    & 46.6 & 57.6 & 64.6 & 69.3 & 70.3 & 74.1 
    & 46.3 & 56.3 & 62.8 & 65.8 & 69.1 & 72.1 \\
    \degree{}~\cite{Hsu22degree}& Gen
    & 41.4 & 56.8 & 62.5 & 68.9 & 70.5 & 73.8
    & 41.7 & 57.7 & 58.9 & 65.8 & 68.2 & 73.0 \\
    \midrule
    \modelbart{} & Gen
    & \ul{52.3} & \textbf{61.5} & \textbf{67.2} & \ul{71.2} & \textbf{72.7} & \ul{75.5}
    & \ul{52.4} & \ul{61.0} & \textbf{66.4} & \textbf{69.7} & \ul{71.1} & \ul{73.4} \\
    \modelroberta{} & Gen
    & \textbf{53.2} & \textbf{61.5} & \ul{66.6} & \textbf{71.8} & \ul{72.5} & \textbf{76.6}
    & \textbf{53.4} & \textbf{61.7} & \textbf{66.4} & \ul{69.5} & \textbf{71.9} & \textbf{74.2} \\
    
    \midrule
    \multicolumn{14}{c}{\textbf{Argument Classification F1-Score (\%) in ERE-EN}} \\
    \midrule
    DyGIE++~\cite{Wadden19dygiepp}& Cls
    & 42.2 & 45.4 & 49.0 & 50.1 & 51.5 & 56.8
    & 40.0 & 44.6 & 49.5 & 52.0 & 53.7 & 56.0 \\
    OneIE~\cite{Lin20oneie}& Cls
    & 51.4 & 59.5 & 62.0 & 65.6 & 68.6 & 71.2 
    & 49.5 & 56.1 & 62.3 & 66.1 & 67.7 & 70.1 \\
    Query and Extract~\cite{wang-etal-2022-query}& Cls
    & 22.0 & 37.3 & 41.2 & 49.4 & 57.0 & 65.0
    & 19.7 & 34.0 & 42.4 & 50.1 & 57.7 & 64.3 \\
    AMR-IE~\cite{DBLP:conf/naacl/ZhangJ21}& Cls
    & 44.8 & 55.2 & 56.8 & 65.2 & 67.6 & 70.1
    & 44.1 & 53.7 & 60.4 & 65.6 & 68.9 & 71.5 \\
    \degree{}~\cite{Hsu22degree}& Gen
    & 57.2 & 62.5 & 63.9 & 67.1 & 70.2 & 73.3
    & 57.5 & 63.9 & 67.4 & 69.1 & \textbf{73.3} & 74.9 \\
    \midrule
    \modelbart{} & Gen
    & \ul{62.4} & \textbf{66.8} & \textbf{66.6} & \ul{68.8} & \textbf{70.8} & \ul{73.6}
    & \ul{62.9} & \ul{66.7} & \textbf{68.5} & \textbf{71.3} & \ul{72.5} & \textbf{75.4}\\
    \modelroberta{} & Gen
    & \textbf{63.1} & \ul{66.7} & \textbf{66.6} & \textbf{69.7} & \ul{70.6} & \textbf{73.8}
    & \textbf{63.2} & \textbf{67.7} & \ul{68.4} & \ul{70.5} & \ul{72.5} & \ul{75.0}\\

    \bottomrule
\end{tabular}
}
\caption{Argument classification F1-scores (\%) under different data proportion settings for ACE05-E and ERE-EN datasets. The highest scores are in bold and the second-best scores are underlined. Generation-based models and Classification-based models are indicated by "Gen" and "Cls" respectively. Due to space constraints, the table with argument identification F1-scores is listed in \Cref{sec:detail_result}.}
\label{table:main_result}
\end{table*}

%% file: 04-experiments.tex
\section{Experiments}
\label{sec:experiment}
We conduct experiments to verify the effectiveness of \model{}. All the reported numbers are the average of the results from three random runs. 

\subsection{Experimental Settings}
\label{subsec:exp_setting}

\paragraph{Datasets and Data split.} 
We adopt the event annotation in ACE 2005 dataset~\cite{doddington-etal-2004-automatic} (\textbf{ACE05-E})\footnote{\url{https://catalog.ldc.upenn.edu/LDC2006T06}}, and the English split in ERE dataset~\cite{DBLP:conf/aclevents/SongBSRMEWKRM15} (\textbf{ERE-EN})\footnote{\url{https://catalog.ldc.upenn.edu/LDC2020T19}}. 
ACE 2005 contains files in English, Chinese, and Arabic, and ERE includes files in English and Chinese.
In this paper, we only use the documents in English, and split them to sentences for use in our experiments.
We follow prior works~\cite{Wadden19dygiepp, Lin20oneie} to preprocess each dataset. After preprocessing, \textbf{ACE05-E} has 33 event types and 22 argument roles, and \textbf{ERE-EN} are with 38 event types and 21 argument roles in total.
Further, we follow \citet{Hsu22degree} to select 5\%, 10\%, 20\%, 30\%, and 50\% of training samples to generate the different data split as the training set for experiments. The data statistics are listed in \Cref{table:dataset_stats} in the Appendix~\ref{sec:dataset}.\footnote{The license for ACE 2005 and ERE is \emph{LDC User Agreement for Non-Members}.
In both datasets, event types like \textit{Conflict:Attack} may include offensive content such as information related to war or violence. Some passages are extracted from broadcast news, thus some real names may appear in the data. Considering the datasets are not publicly available, and these contents are likely to be the arguments to extract in our task, we do not change the data for protecting or anonymizing.}

\paragraph{Evaluation metrics.}
We report the F1-score for argument predictions following prior works~\cite{Wadden19dygiepp, Lin20oneie}. An argument is correctly identified (\textbf{Arg-I}) if the predicted span matches the span of any gold argument; it is correctly classified (\textbf{Arg-C}) if the predicted role type also matches.

\paragraph{Implementation details.}
We use the AMR tools as we mentioned in \Cref{sec:method}. When training our models, we set the learning rate to $10^{-5}$. The number of training epochs is 60 when training on ACE05E, and 75 when training on ERE-EN. We simply set $\lambda$ as 1 for all our models. We do hyper-parameter searching using the setting that trains on 20\% of data in ACE05E and selects the best model based on the development set results. We set $l=40$, and batch size is set to 4 for \modelbart{} and 6 for \modelroberta{} in the end. This is searching from $l=\{30, 40, 50\}$ and batch size $=\{4, 6, 8, 12\}$.

\paragraph{Baselines.} 
We compare \model{} with the following classification-based models: 
(1) \textbf{DyGIE++}~\cite{Wadden19dygiepp}, which extracts information by scoring spans with contextualized representations.
(2) \textbf{OneIE}~\cite{Lin20oneie}, a joint IE framework that incorporates global features.
(3) \textbf{Query and Extract}~\cite{wang-etal-2022-query}, which uses attention mechanisms to evaluate the correlation between role names and candidate entities.
(4) \textbf{AMR-IE}~\cite{DBLP:conf/naacl/ZhangJ21}, which captures non-local connections between entities by aggregating neighborhood information on AMR graph, and designed hierarchical decoding based on AMR graph information.
We also consider the following generation-based models:
(5) \textbf{PAIE}~\cite{ma-etal-2022-prompt}, a framework that integrated prompt tuning, and generates span selectors for each role. \footnote{PAIE requires manually designed prompts for model training, and ERE-EN dataset is not considered in their official codebase. Hence, we do not include PAIE as a baseline on ERE-EN.}
(6) \textbf{\degree{}}~\cite{Hsu22degree}. The generation-based EAE model we used as our base model.

To ensure a fair comparison across models, we adopt the official codes of the above baselines to train them on the identical data and did hyper-parameter tuning. For all the classification-based methods, we use \texttt{RoBERTa-large}, and for all the generation-based methods, we use \texttt{BART-large} as the pre-trained language models. \Cref{subsec:baselines} shows details about the implementation.

\begin{table*}[t!]
\centering
\small
\resizebox{1.0\textwidth}{!}{
\setlength{\tabcolsep}{3.5pt}
\renewcommand{\arraystretch}{1.05}
\begin{tabular}{lcccccccc}
\toprule
\multirow{3}{*}{Model}       & \multicolumn{4}{c}{5\% ACE05-E Data} & \multicolumn{4}{c}{20\% ACE05-E Data} \\
\cmidrule{2-9}
& \multicolumn{2}{c|}{Dev. Set} & \multicolumn{2}{c}{Test Set}
& \multicolumn{2}{c|}{Dev. Set} & \multicolumn{2}{c}{Test Set} \\
\cmidrule{2-9}
& Arg-I         & Arg-C         & Arg-I          & Arg-C
& Arg-I         & Arg-C         & Arg-I          & Arg-C \\
\midrule
\model{} w/o \AMRPrefix{}  & 57.8{\tiny$\pm 2.32$} & 49.8{\tiny$\pm 2.59$} & 55.9{\tiny$\pm 0.75$} & 47.5{\tiny$\pm 0.32$} & 70.6{\tiny$\pm 0.94$} & 64.5{\tiny$\pm 0.30$} & 67.0{\tiny$\pm 0.81$} & 62.6{\tiny$\pm 1.36$} \\ 
\midrule
\modelbart{}     & 59.9{\tiny$\pm 1.99$} & 52.3{\tiny$\pm 1.54$} & 59.8{\tiny$\pm 2.00$} & 52.4{\tiny$\pm 1.53$} & 72.0{\tiny$\pm 0.80$} & \textbf{67.2}{\tiny$\pm 0.55$} & 70.2{\tiny$\pm 0.84$} & \textbf{66.4}{\tiny$\pm 1.04$} \\
\modelroberta{}  & 62.1{\tiny$\pm 1.73$} & \textbf{53.2}{\tiny$\pm 2.26$} & \textbf{61.0}{\tiny$\pm 0.98$} & \textbf{53.4}{\tiny$\pm 0.21$} & 71.5{\tiny$\pm 1.00$} & 66.6{\tiny$\pm 0.12$} & \textbf{70.5}{\tiny$\pm 1.28$} & \textbf{66.4}{\tiny$\pm 0.86$} \\  
\modelbart{} w/ frozen AMR Encoder    & 60.9{\tiny$\pm 2.10$} & 51.5{\tiny$\pm 1.78$} & 58.3{\tiny$\pm 1.63$} & 51.1{\tiny$\pm 1.21$} & \textbf{72.5}{\tiny$\pm 0.50$} & 66.5{\tiny$\pm 1.06$} & 70.0{\tiny$\pm 0.37$} & 65.8{\tiny$\pm 0.19$} \\
\modelroberta{} w/ frozen AMR Encoder    & \textbf{62.5}{\tiny$\pm 1.49$} & 50.9{\tiny$\pm 1.34$} & 60.6{\tiny$\pm 0.46$} & 50.7{\tiny$\pm 0.09$} & 71.7{\tiny$\pm 0.50$} & 66.0{\tiny$\pm 0.76$} & 69.8{\tiny$\pm 1.52$} & 65.5{\tiny$\pm 1.47$} \\ 
AMR Prompts     & 56.7{\tiny$\pm 1.00$} & 48.4{\tiny$\pm 1.11$} & 55.2{\tiny$\pm 1.33$} & 47.2{\tiny$\pm 1.25$} & 71.2{\tiny$\pm 0.66$} & 65.7{\tiny$\pm 0.80$} & 69.5{\tiny$\pm 0.26$} & 64.9{\tiny$\pm 0.51$} \\
AMRBART Encoding Concatenation & 58.4{\tiny$\pm 0.45$} & 50.3{\tiny$\pm 1.01$} & 56.4{\tiny$\pm 2.16$} & 48.3{\tiny$\pm 1.50$} & 71.2{\tiny$\pm 0.87$} & 64.7{\tiny$\pm 0.14$} & 69.1{\tiny$\pm 1.54$} & 64.4{\tiny$\pm 1.33$} \\
RoBERTa Encoding Concatenation  & 6.4{\tiny$\pm 0.85$} & 4.8{\tiny$\pm 1.10$} & 4.9{\tiny$\pm 2.81$} & 3.3{\tiny$\pm 1.96$} & 11.6{\tiny$\pm 2.58$} & 8.7{\tiny$\pm 0.55$} & 11.4{\tiny$\pm 1.38$} & 10.4{\tiny$\pm 1.69$} \\
\bottomrule

\end{tabular}
}
\caption{Ablation study of different ways for AMR incorporation. Report numbers in F1-scores (\%).}
\label{table:ablation_amr_detail}
\end{table*}

\subsection{Results}
\label{subsec:results}
\Cref{table:main_result} shows the argument classification (\textbf{Arg-C}) F1-scores in ACE05-E and ERE datasets under different data proportions. 
Overall, both \modelroberta{} and \modelbart{} consistently outperform all other baselines except the test set results of using 50\% data in ERE-EN.

From the table, we can notice that \model{} significantly outperforms our base model DEGREE in all experiments in ACE05-E, and in ERE-EN, the improvement is also considerable. When trained with less than 20\% data in ACE05-E, \modelroberta{} can consistently achieve more than 4 points of improvement over \degree{} in both the development and test sets. In the following \Cref{sec:ablation}, we will further discuss the detailed contribution of our method over \degree{}.

To quantitatively evaluate the effectiveness of AMR's incorporation, we can first check the performance of AMR-IE. 
AMR-IE achieves competitive performance among classification-based models, especially under extremely low-resource settings. 
This is coincident with how \model{}'s result shows.
\model{} outperforms both \degree{} and PAIE, and the gap is more obvious under low-resource settings. For example, in the 5\% data proportion setting, \modelroberta{} made over 11 points of improvement over \degree{} in ACE05-E's test Set. In the meanwhile, \modelroberta{} achieves 4.4 points of performance gain compared with PAIE.
All this shows the empirical evidence that AMR information can hint to the models' semantic structure of the input passage, and this is especially helpful for models when training samples are limited.
Despite the strong performance of AMR-IE, \model{} can still outperform it across all the settings, indicating the effectiveness of our method.

Comparing \modelbart{} and \modelroberta{}, we show that our proposed method does not necessarily rely on pre-trained AMR-to-Text models. Particularly, \modelroberta{}, which employs a pre-trained \texttt{RoBERTa-large} as the AMR Encoder still achieves competitive results to \modelbart{}, which uses AMR-to-Text data. Yet, the advantage of using AMR-to-Text data as an auxiliary is that we can get similar results with less parameters. The \textit{AMR encoder component} of \modelroberta{} has approximately 1.7 times more parameters than that of \modelbart{}, as we only use the encoder part of \texttt{AMRBART} in AMPERE(AMRBART). Nevertheless, the pre-trained knowledge from AMR-to-text data enables \modelbart{} to perform competitively with \modelroberta{}.

%% file: 05-studies.tex
\section{Analysis}
\label{sec:ablation}

In this section, we present comprehensive ablation studies and case studies to validate our model designs. 
Two essential parts of our design, the AMR-aware prefix, and the adjusted copy mechanism will be examined in the following studies. For all the experiments in this section, we use the setting of training on 5\% and 20\% ACE05-E data to simulate very low-resource and low-resource settings.

\subsection{Different Ways for AMR Incorporation}

We compare different ways to incorporate AMR information into generation-based EAE models:
\begin{itemize} [topsep=3pt, itemsep=-2pt, leftmargin=13pt]
\item \textbf{AMR Prompts.} We follow the same process as \model{} to obtain the linearized AMR graph sequence. Then, we directly concatenate the linearized AMR graph sequence to the input text as part of the prompts.
\item \textbf{\texttt{AMRBART} Encoding Concatenation.} After obtaining the AMR sequence representations after the AMR encoder using \texttt{AMRBART}, we concatenate this encoding with the output representation in our generation-based EAE model and feed them together to the decoder.
\item \textbf{\texttt{RoBERTa} Encoding Concatenation.} The method is similar to the AMRBART Encoding Concatenation method, except that we use \texttt{RoBERTa} as the AMR encoder.~\footnote{We also explore the variation that we add a linear layer to the AMR encoding to help space alignment, but there is little performance difference on both \texttt{AMRBART} Encoding Concatenation \& \texttt{RoBERTa} Encoding Concatenation.}
\end{itemize}
For comparison, we provide \model{}'s performance without any AMR incorporation as a baseline. 
Additionally, we also consider \model{} with frozen AMR encoder\footnote{For these type of models, during training, the AMR encoder's parameter is fixed. Hence the number of learnable parameters is comparable to \degree{}.} in the comparisons to exclude the concern of extra learnable parameters of \model{} compared to baselines such as AMR Prompts.
Note that all the mentioned models above are implemented with our proposed adjusted copy mechanism.
\Cref{table:ablation_amr_detail} shows the results.

From the table, we observe that \model{} gets the best performance among all the ways we explored and achieves 4.9\% and 4.2\% F1-score improvements over the model without AMR incorporation under the case of using 5\% \& 20\% of training data, respectively.

An interesting finding is that the performance of AMR Prompts is worse than the model without any AMR incorporation in the very low-resource setting (5\% data). As mentioned in \Cref{sec:intro}, the heterogeneous nature between AMR graph and natural language sentences is an important intuitive for our model design. AMR often uses special tokens such as \textit{:ARG0} or \textit{appeal-01}, and in implementation like AMR Prompts, it would be confusing for models when training samples are not sufficient.

Furthermore, due to the heterogeneous vector space between \texttt{AMRBART} and \texttt{RoBERTa}, \texttt{RoBERTa} Encoding Concatenation method could not work well. In comparison, the prefix design of \model{} shows strong adaptability, as \modelbart{} and \modelroberta{} both outperform the other implementation methods.

Finally, we focus on the results from \model{} with frozen AMR Encoder. We can observe that despite slight performance degradation compared to fully-trainable \model{}, \model{} with frozen AMR Encoder still obtain at least 1\% absolute F1-scores improvements over other AMR incorporation methods.

\begin{table*}[t!]
\centering
\small
\setlength{\tabcolsep}{3.5pt}
\renewcommand{\arraystretch}{1.05}
\resizebox{1.0\textwidth}{!}{
\begin{tabular}{llcccccccc}
\toprule
\multicolumn{2}{l}{\multirow{3}{*}{Model}}       & \multicolumn{4}{c}{5\% ACE05-E Data} & \multicolumn{4}{c}{20\% ACE05-E Data} \\
\cmidrule{3-10}
&& \multicolumn{2}{c|}{Dev. Set} & \multicolumn{2}{c}{Test Set}
& \multicolumn{2}{c|}{Dev. Set} & \multicolumn{2}{c}{Test Set} \\
\cmidrule{3-10}
&& Arg-I         & Arg-C         & Arg-I          & Arg-C
& Arg-I         & Arg-C         & Arg-I          & Arg-C \\
\midrule
\modelbart{}  & w/ adjusted copy mechanism    & \textbf{59.9}{\tiny$\pm 1.99$} & \textbf{52.3}{\tiny$\pm 1.54$} & \textbf{59.8}{\tiny$\pm 2.00$} & \textbf{52.4}{\tiny$\pm 1.53$} & \textbf{72.0}{\tiny$\pm 0.80$} & \textbf{67.2}{\tiny$\pm 0.55$} & 70.2{\tiny$\pm 0.84$} & \textbf{66.4}{\tiny$\pm 1.04$}   \\
  & w/o \anycopy{}     & 48.7{\tiny$\pm 0.67$} & 41.3{\tiny$\pm 1.56$} & 49.0{\tiny$\pm 1.68$} & 43.2{\tiny$\pm 0.77$} & 69.8{\tiny$\pm 0.64$} & 63.7{\tiny$\pm 0.27$} & 65.5{\tiny$\pm 0.90$} & 61.4{\tiny$\pm 0.93$} \\
 & w/ \purecopy{}   & 57.0{\tiny$\pm 1.25$} & 50.4{\tiny$\pm 2.76$} & 55.0{\tiny$\pm 1.59$} & 49.0{\tiny$\pm 0.84$} & 70.0{\tiny$\pm 0.89$} & 65.6{\tiny$\pm 0.80$} & 67.3{\tiny$\pm 1.77$} & 62.8{\tiny$\pm 1.79$} \\
 & w/ \noregcopy{}   & 58.9{\tiny$\pm 1.28$} & 49.9{\tiny$\pm 0.32$} & 53.4{\tiny$\pm 3.34$} & 48.1{\tiny$\pm 2.61$} & 70.7{\tiny$\pm 0.58$} & 64.1{\tiny$\pm 0.26$} & 68.3{\tiny$\pm 1.07$} & 63.6{\tiny$\pm 0.30$} \\
\midrule
\modelroberta{} & w/ adjusted copy mechanism   & \textbf{62.1}{\tiny$\pm 1.73$} & \textbf{53.2}{\tiny$\pm 2.26$} & \textbf{61.0}{\tiny$\pm 0.98$} & \textbf{53.4}{\tiny$\pm 0.21$} & 71.5{\tiny$\pm 1.00$} & \textbf{66.6}{\tiny$\pm 0.12$} & \textbf{70.5}{\tiny$\pm 1.28$} & \textbf{66.4}{\tiny$\pm 0.86$} \\
& w/o \anycopy{}    & 52.5{\tiny$\pm 0.85$} & 44.0{\tiny$\pm 1.22$} & 50.7{\tiny$\pm 1.79$} & 44.5{\tiny$\pm 1.81$} & 69.9{\tiny$\pm 0.16$} & 62.7{\tiny$\pm 0.26$} & 65.5{\tiny$\pm 1.01$} & 61.1{\tiny$\pm 1.34$} \\
 & w/ \purecopy{}   & 56.9{\tiny$\pm 1.63$} & 48.5{\tiny$\pm 1.08$} & 55.5{\tiny$\pm 1.62$} & 48.6{\tiny$\pm 0.50$} & 71.1{\tiny$\pm 0.67$} & 66.3{\tiny$\pm 1.20$} & 67.3{\tiny$\pm 1.32$} & 63.7{\tiny$\pm 1.28$} \\
 & w/ \noregcopy{}  & 58.5{\tiny$\pm 2.03$} & 50.7{\tiny$\pm 0.18$} & 54.5{\tiny$\pm 1.01$} & 47.8{\tiny$\pm 1.09$} & 71.1{\tiny$\pm 1.40$} & 64.9{\tiny$\pm 0.74$} & 69.0{\tiny$\pm 1.61$} & 64.3{\tiny$\pm 1.16$} \\
\bottomrule

\end{tabular}
}
\caption{The study of using different generation mechanisms. Report numbers in F1-scores (\%). The best performance among methods using the same model architecture is highlighted in bold.}
\label{table:ablation_copy_detail}
\end{table*}

\subsection{Studies of Adjusted Copy Mechanism}

To justify the effectiveness of our adjusted copy mechanism, we compare our adjusted copy mechanism with the following method:
\begin{itemize}[topsep=3pt, itemsep=-2pt, leftmargin=13pt]
\item \textbf{\model{} w/o \anycopy{}.} For comparison, we adopt a normal generation-based model adapted with AMR-aware prefixes.
\item \textbf{\model{} w/ \purecopy{}.}: In \Cref{equation:copy}, we directly set $w^i_{gen}=0$. In other words, tokens not presented in the input can not be generated.
\item \textbf{\model{} w/ \noregcopy{}.} We apply the copy mechanism but train the model with the loss function in \Cref{equation:normal-loss}. 
\end{itemize}

In \Cref{table:ablation_copy_detail}, the experiment with \modelbart{} and \modelroberta{} lead to similar conclusions. 
Any kind of copy mechanism can lead to noticeable improvement, and the performance gap between methods with and without copy mechanism is larger in the lower data proportion setting.
Our adjusted copy mechanism stably outperforms the other methods in studies.
Compared to the traditional copy mechanism, our method encourages the model to copy more, hence can stably overcome the very low-resource challenges.
Compared to fully relying on copy from input, our method allows the generative model to freely generate tokens not presented in the input, so as to better leverage the pre-trained language model's power, leading to better performance when data is slightly more available.


\subsection{Case Study}
\subsubsection{Output Examples}
\begin{figure*}[t!]
\centering 
\includegraphics[width=0.98\textwidth]{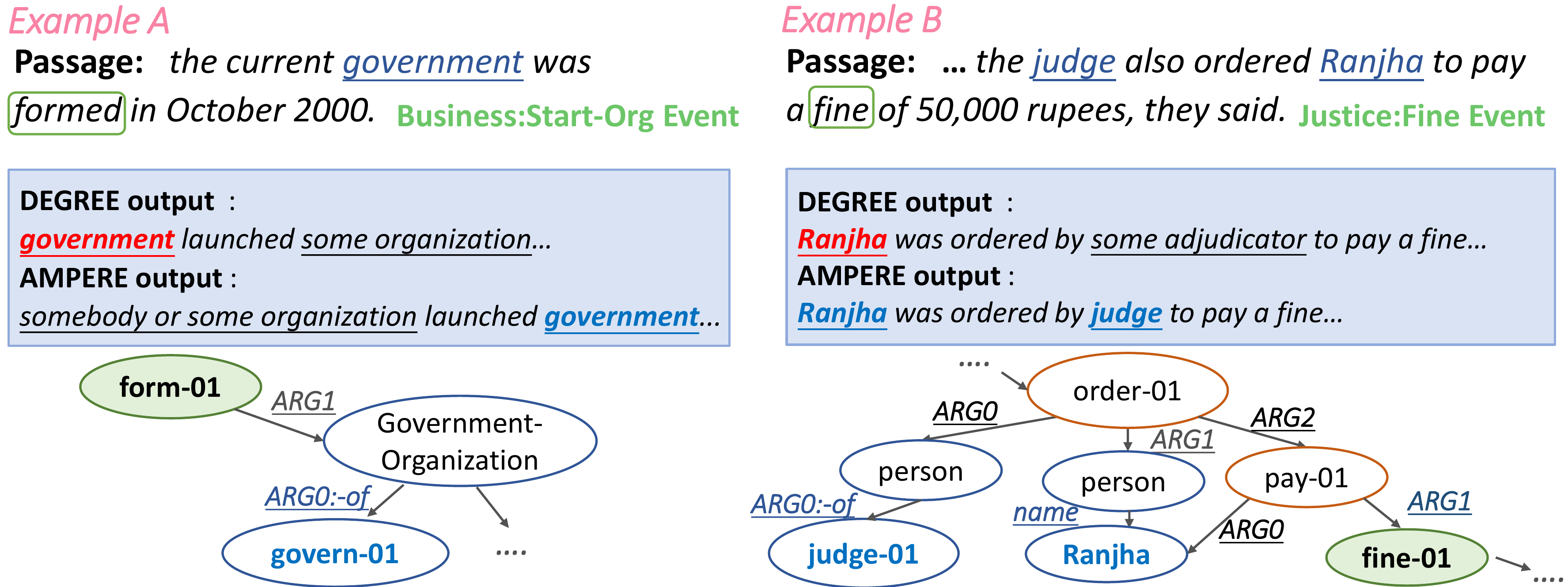} 
\caption{Two examples of how AMR information helps the generation of event argument predictions. Note that due to space constraints, the shown passage, output text, and AMR graph omit some irrelevant information. } 
\label{fig:case_study} 
\end{figure*}

To intuitively explain the benefit of our method over previous generation-based EAE models, we present examples here to showcase the influence of incorporating AMR information.
We compare \model{} and \degree{} that both trained on 20\% ACE05-E data and demonstrate two examples in \Cref{fig:case_study} to show the difference of their generated output text.

\paragraph{Example A} presents a case where the edges in the AMR graph helps the model to classify the correct role type of argument \textit{"government"}.
Without AMR information, \degree{} incorrectly predicts the \textit{"government"} to be the agent that launched some organization. 
In the AMR graph, edge \textit{ARG1} points to the object of the action concept \textit{form-01}.
Thus, in the \model{}'s output, \textit{"government"} is correctly classified as the object of \textit{"form"}.

\paragraph{Example B} in \Cref{fig:case_study} shows how the AMR graph hints \model{} about the argument \textit{"judge"}.
By looking up the subject of verb \textit{"order"} in the AMR graph, the model is able to find the adjudicator of the event. Thus, \model{} could correctly replace the adjudicator placeholder in the template with real adjudicator, \textit{``judge''}.

\subsubsection{Error Analysis}
To point out future research direction for generation-based EAE models, we performed error analysis on 30 cases where our \modelroberta{} made mistakes. We identified two common types of errors: (1) ambiguous span boundaries, and (2) incorrect distinction between events of the same type. 

For instance, in the case of ``ambiguous span boundaries,'' \modelroberta{} incorrectly predicted \textit{"Christian Ayub Masih"} instead of the correct label \textit{"Ayub Masih."} We observe that generation-based models struggle to accurately predict span boundaries, as both \modelroberta{}'s output and the ground truth can fit into the sentence template coherently. Even with the inclusion of AMR, the model's ability to identify potential boundaries from the AMR graph through learning remains limited.

Regarding the issue of ``incorrect distinction between events of the same type,'' we present an example to illustrate this. In the given input sentence, \textit{``As well as previously holding senior positions at Barclays Bank, BZW and Kleinwort Benson, McCarthy was formerly a top civil servant at the Department of Trade and Industry.''}, the model becomes confused between the two "Personnel:End-Position" events, each triggered by \textit{``previousl''} and \textit{``formerly''}, respectively, due to subtle differences. We suggest that incorporating additional structural knowledge, such as dependency parsing information, to separate the sentences structurally, could be a potential solution. However, we leave this research as future works.


%% file: 02-related.tex
\section{Related Work}
\label{sec:related}
\paragraph{Generation-based event (argument) extraction models.}
Traditionally, most models for EAE are classification-based~\cite{DBLP:conf/acl/ChenXLZ015, DBLP:conf/emnlp/MaWABA20, DBLP:journals/corr/abs-2205-12585, DBLP:conf/aaai/FinckeAMB22}. Recently, generation-based EAE models~\cite{Hsu22degree, DBLP:conf/acl/0001LXHTL0LC20, DBLP:conf/naacl/LiJH21, DBLP:conf/iclr/PaoliniAKMAASXS21} become more and more popular due to their flexibility to present different output structures~\cite{DBLP:conf/acl/YanGDGZQ20}, to be unified considered with similar tasks~\cite{DBLP:conf/acl/0001LDXLHSW22}, and their competitive performance~\cite{Hsu22degree, DBLP:conf/acl/LiuHSW22}. 

The development of generation-based event (argument) extraction models starts from works investigating how to reformulate event extraction problems as a generation task~\cite{DBLP:conf/eacl/DuRC21, DBLP:conf/naacl/DuRC21}. Follow-up works put efforts to show the influence of different prompt designs to the generative event models.~\cite{ma-etal-2022-prompt, DBLP:journals/corr/abs-2210-10709, Hsu22degree} 
More recently, researchers start to improve this series of work by designing different model architectures~\cite{DBLP:conf/acl/DuLJ22, DBLP:conf/ijcai/0002QCWHY022}.
However, very few efforts have been put into the ways and the effectiveness of incorporating auxiliary syntactic and semantic information into such models, even though this information has been shown to be beneficial in classification-based models. Hence, in this paper, we present the study and explore ways to incorporate this additional information for generation-based event models.

\paragraph{Improving event extraction with weakly-supervisions.}
Being a challenging task that requires deep natural language understanding to solve, many prior efforts have been put into investigating which auxiliary upstream task information is useful for event predictions.~\cite{DBLP:journals/corr/abs-2205-12490, DBLP:conf/emnlp/LiuCLZ19, DBLP:conf/acl/DaganJVHCR18,  DBLP:conf/emnlp/VeysehNN20} \citet{DBLP:conf/emnlp/LiuCLZ19, DBLP:conf/aaai/AhmadPC21} leverages dependency syntactic structures of the input sentence to help cross-lingual event predictions. \citet{DBLP:conf/acl/HuangCFJVHS16, DBLP:conf/acl/DaganJVHCR18} uses the similarity between AMR and event structures to perform zero-shot event extraction. More recently, \citet{DBLP:conf/naacl/ZhangJ21, DBLP:conf/emnlp/VeysehNN20, DBLP:conf/naacl/XuWLZCS22} investigates different message passing methods on AMR graph to help learn better representations for final classifications. Despite many efforts that have been put into the community, these methods are designed for classification-based models. 
This highlights the open area for research --- how and whether incorporating such auxiliary information can also be helpful. We take a step forward in this direction and present \model{} to showcase the possibility to improve the generation-based event models by such way.

%% file: 06-conclusion.tex
\section{Conclusion}
\label{sec:conclusion}
In this paper, we present \model{}, a generation-based model equipped with AMR-aware prefixes. 
Through our comprehensive studies, we show that prefixes can serve as an effective medium to connect AMR information and the space of generative models, hence achieving effective integration of the auxiliary semantic information to the model. Additionally, we introduce an adjusted copy mechanism to help \model{} more accurately and stably generate output disregarding the additional noise brought from the AMR graph.
Through our experiments, we show that \model{} achieves consistent improvements in every setting, and the improvement is particularly obvious in low-resource settings.
\looseness=-1

\section*{Acknowledgments}
We thank anonymous reviewers for their helpful feedback. 
We thank the UCLA PLUSLab and UCLA-NLP group members for their initial review and feedback for an earlier version of the paper. 
This research was supported in part by AFOSR MURI via Grant \#FA9550-22-1-0380, Defense Advanced Research Project Agency (DARPA) via Grant \#HR00112290103/HR0011260656, the Intelligence Advanced Research Projects Activity (IARPA) via Contract No. 2019-19051600007, National Science Foundation (NSF) via Award No.~2200274, and a research award sponsored by CISCO.

\section*{Limitations}
Our goal is to demonstrate the potential of incorporating AMR to improve generation-based EAE models.
Although we have shown the strength of our method, there are still some limitations.
First, our proposed techniques are based on the AMR graph generated by pre-trained AMR parsers. The generated AMR graphs inevitably have a certain possibility of being not perfect. Hence, the error propagation issues would happen to \model{}. We hypothesize this is one of the reasons why the improvement of \model{} is not necessarily significant when data is abundant. Yet, through our experimental results, we still show the benefit of incorporating this information, especially in the case of low-resource settings.
Second, although our AMR-aware prefix design should be agnostic to the used generation-based EAE model, in our experiment, we only set \degree{} as our base generation-based EAE model. We leave the investigation on the generalizability of our AMR-prefix method to other base models as future work.

\section*{Ethics Considerations}
Our method relies on a pre-trained AMR parser, which is built using pre-trained large language models (\texttt{AMRBART} \& \texttt{RoBERTa}). It is known that the models trained with a large text corpus may capture the bias reflecting the training data. 
Therefore, it is possible that the AMR graph used in our method could contain certain biases.
We suggest carefully examining the potential bias before applying \model{} to any real-world applications.

%% file: 99-appendix.tex
\clearpage
\appendix

\section{Datasets}
\label{sec:dataset}
We present detailed dataset statistics in Table \Cref{table:dataset_stats}.


\begin{table*}[t!]
\centering
\small
\setlength{\tabcolsep}{4.5pt}
\begin{tabular}{l|c|cccccc}
    \toprule
    Dataset & Split & \#Docs& \#Sents & \#Events & \#Event Types & \#Args & \#Arg Types\\
    \midrule
    \multirow{8}{*}{ACE05-E}
    & Train (5\%) & 25& 649& 212& 27& 228& 21 \\
    & Train (10\%) & 50& 1688& 412& 28& 461& 21 \\
    & Train (20\%) &110& 3467& 823& 33& 936& 22 \\
    & Train (30\%) &160& 5429& 1368& 33& 1621& 22 \\
    & Train (50\%) &260& 8985& 2114& 33& 2426& 22 \\
    & Train (full) & 529 & 17172 & 4202 & 33 & 4859 & 22 \\
    \cmidrule{2-8}
    & Dev   &   28& 923& 450& 21& 605& 22 \\
    & Test  &   40& 832& 403& 31& 576& 20 \\
    \midrule
    \multirow{8}{*}{ERE-EN}
    & Train (5\%) & 20& 701& 437& 31& 640& 21 \\
    & Train (10\%) & 40& 1536& 618& 37& 908& 21 \\
    & Train (20\%) & 80& 2848& 1231& 38& 1656& 21 \\
    & Train (30\%) &120& 4382& 1843& 38& 2632& 21 \\
    & Train (50\%) &200& 7690& 3138& 38& 4441& 21 \\
    & Train (full) &396& 14736& 6208& 38& 8924& 21 \\
    \cmidrule{2-8}
    & Dev   &   31& 1209& 525& 34& 730& 21 \\
    & Test  &   31& 1163& 551& 33& 822& 21 \\
    \bottomrule

\end{tabular}
\caption{Dataset statistics. ``\#Docs'' means the number of documents; ``\#Sents'' means the number of sentences, ``\#Events'' means the number of events in total; ``\#Event Types'' means the size of event types set; ``\#Args'' means the number of argument in total; ``\#Arg Types'' means the size of argument role types set.}
\label{table:dataset_stats}
\end{table*}

\section{Implementation Details}

\label{subsec:baselines}
This section introduces the implementation details for all the baseline models we use in this paper. Our experiments are run using our machine that equips 8 NVIDIA a6000 GPUs.

\begin{itemize}
\item \textbf{DyGIE++}: we use their official code to reimplement the model.\footnote{\url{https://github.com/dwadden/dygiepp}} Their original model is built using \texttt{BERT}~\cite{DBLP:conf/naacl/DevlinCLT19}. As we mentioned in \Cref{subsec:exp_setting}, we replace the used pre-trained language model into \texttt{RoBERTa-large} and tune with learning rates $= \{1e-5, 2e-5, 3e-5\}$.
\item \textbf{OneIE}: we use their official code\footnote{\url{http://blender.cs.illinois.edu/software/oneie/}} to train the model. Their original model is built using \texttt{BERT}~\cite{DBLP:conf/naacl/DevlinCLT19}. As we mentioned in \Cref{subsec:exp_setting}, we replace the used pre-trained language model into \texttt{RoBERTa-large} and tune with learning rates $= \{1e-5, 2e-5, 3e-5\}$.
\item \textbf{Query and Extract}: we use their official code\footnote{\url{https://github.com/VT-NLP/Event_Query_Extract/}} to train argument detection model with learning rate $= 1e-5$, batch size $= 16$, training epoch $=10$. Different from the official code, we used \texttt{RoBERTa-large} for pre-trained language model to ensure a fair comparison. 
\item \textbf{AMR-IE}: the original AMR-IE is an end-to-end event extraction model, so we adapt their official code\footnote{\url{https://github.com/zhangzx-uiuc/AMR-IE}} to event argument extraction task by giving gold triggers in model evaluation. We fixed pre-trained language model learning rate $=1e-5$, then did hyperparameter searching from graph learning rate $= \{1e-3,4e-3\}$ and batch size $= \{8,16\}.$
\item \textbf{PAIE}: we use their official code\footnote{\url{https://github.com/mayubo2333/PAIE/}} to train the model with the default parameters for \texttt{BART-large}.
\item \textbf{DEGREE}: we use their official code\footnote{\url{https://github.com/PlusLabNLP/DEGREE}} to train the model with the default parameters for \texttt{BART-large}.
\end{itemize}

\section{Detailed Result}
\label{sec:detail_result}
\Cref{table:main_result_detail} shows the detailed results of our main experiments. We repeat running every experiment setting with three random seeds, and report their average Arg-I and Arg-C F1-scores, and the corresponding standard deviation scores.

\begin{table*}[t!]
\centering
\small
\resizebox{1.0\textwidth}{!}{
\setlength{\tabcolsep}{3.5pt}
\renewcommand{\arraystretch}{1.05}
\begin{tabular}{lccccccccccccc}
    \toprule
    \multicolumn{13}{c}{\textbf{ACE05-E Development Set}} \\
    \midrule
    \multirow{2}{*}{Model} & \multicolumn{2}{c|}{5\%} & \multicolumn{2}{c|}{10\%} & \multicolumn{2}{c|}{20\%} & \multicolumn{2}{c|}{30\%} & \multicolumn{2}{c|}{50\%} & \multicolumn{2}{c}{100\%} \\
    \cmidrule{2-13}
    & Arg-I     & Arg-C     & Arg-I     & Arg-C     & Arg-I     & Arg-C     & Arg-I     & Arg-C     & Arg-I     & Arg-C     & Arg-I     & Arg-C \\
    \midrule
    DyGIE++~\cite{Wadden19dygiepp}
    & 44.6{\tiny$\pm 2.28$} & 34.6{\tiny$\pm 1.83$} & 57.3{\tiny$\pm 0.91$} & 48.5{\tiny$\pm 0.35$} & 58.9{\tiny$\pm 1.53$} & 52.5{\tiny$\pm 0.85$} & 63.0{\tiny$\pm 2.05$} & 57.5{\tiny$\pm 1.34$} & 65.4{\tiny$\pm 0.49$} & 57.9{\tiny$\pm 0.59$} & 67.2{\tiny$\pm 1.78$} & 60.0{\tiny$\pm 0.35$} \\
    OneIE~\cite{Lin20oneie}
    & 48.0{\tiny$\pm 2.27$} & 38.6{\tiny$\pm 1.11$} & 62.3{\tiny$\pm 0.61$} & 56.0{\tiny$\pm 1.01$} & 68.2{\tiny$\pm 0.84$} & 63.2{\tiny$\pm 1.16$} & 73.0{\tiny$\pm 1.20$} & 67.6{\tiny$\pm 0.42$} & 74.6{\tiny$\pm 0.60$} & 70.4{\tiny$\pm 0.46$} & 76.0{\tiny$\pm 1.95$} & 71.8{\tiny$\pm 1.54$} \\
    Query and Extract~\cite{wang-etal-2022-query}
    & 41.6{\tiny$\pm 2.50$} & 10.5{\tiny$\pm 0.82$} & 43.0{\tiny$\pm 2.10$} & 27.7{\tiny$\pm 1.00$} & 49.0{\tiny$\pm 3.00$} & 37.6{\tiny$\pm 0.66$} & 58.8{\tiny$\pm 1.83$} & 50.0{\tiny$\pm 1.52$} & 61.7{\tiny$\pm 3.33$} & 54.6{\tiny$\pm 3.58$} & 67.9{\tiny$\pm 1.86$} & 61.7{\tiny$\pm 2.67$} \\
    AMR-IE~\cite{DBLP:conf/naacl/ZhangJ21}
    & 49.7{\tiny$\pm 1.12$} & 40.0{\tiny$\pm 1.29$} & 62.0{\tiny$\pm 0.34$} & 56.4{\tiny$\pm 0.83$} & 66.8{\tiny$\pm 0.90$} & 61.3{\tiny$\pm 1.23$} & 72.4{\tiny$\pm 1.28$} & 67.4{\tiny$\pm 0.66$} & 74.7{\tiny$\pm 1.04$} & 70.6{\tiny$\pm 1.30$} & 77.7{\tiny$\pm 0.93$} & 73.1{\tiny$\pm 0.68$} \\
    PAIE~\cite{ma-etal-2022-prompt}
     & 55.2{\tiny$\pm 1.16$} & 46.6{\tiny$\pm 0.98$} & 64.1{\tiny$\pm 0.88$} & 57.6{\tiny$\pm 1.43$} & 70.4{\tiny$\pm 0.49$} & 64.6{\tiny$\pm 1.13$} & 74.5{\tiny$\pm 0.38$} & 69.3{\tiny$\pm 0.38$} & 75.1{\tiny$\pm 1.89$} & 70.3{\tiny$\pm 1.02$} & 78.5{\tiny$\pm 0.65$} & 74.1{\tiny$\pm 0.80$} \\
    \degree{}~\cite{Hsu22degree}
    & 47.6{\tiny$\pm 0.64$} & 41.4{\tiny$\pm 0.50$} & 65.1{\tiny$\pm 0.75$} & 56.8{\tiny$\pm 0.50$} & 69.7{\tiny$\pm 0.50$} & 62.5{\tiny$\pm 0.55$} & 75.6{\tiny$\pm 0.43$} & 68.9{\tiny$\pm 0.54$} & 75.9{\tiny$\pm 0.57$} & 70.5{\tiny$\pm 0.28$} & 78.4{\tiny$\pm 0.38$} & 73.8{\tiny$\pm 0.58$} \\
    \midrule
    \modelbart{}
     & \ul{59.9}{\tiny$\pm 1.99$} & \ul{52.3}{\tiny$\pm 1.54$} & \textbf{68.5}{\tiny$\pm 0.83$} & \textbf{61.5}{\tiny$\pm 0.82$} & \textbf{72.0}{\tiny$\pm 0.80$} & \textbf{67.2}{\tiny$\pm 0.55$} & \ul{76.5}{\tiny$\pm 1.01$} & \ul{71.2}{\tiny$\pm 0.56$} & \textbf{76.5}{\tiny$\pm 0.50$} & \textbf{72.7}{\tiny$\pm 0.83$} & \ul{80.0}{\tiny$\pm 1.06$} & \ul{75.6}{\tiny$\pm 1.10$} \\
    \modelroberta{} 
    & \textbf{62.1}{\tiny$\pm 1.73$} & \textbf{53.2}{\tiny$\pm 2.26$} & \ul{68.2}{\tiny$\pm 0.39$} & \textbf{61.5}{\tiny$\pm 1.24$} & \ul{71.5}{\tiny$\pm 1.00$} & \ul{66.6}{\tiny$\pm 0.12$} & \textbf{76.8}{\tiny$\pm 0.37$} & \textbf{71.8}{\tiny$\pm 0.53$} & \ul{76.4}{\tiny$\pm 1.01$} & \ul{72.5}{\tiny$\pm 0.79$} & \textbf{80.9}{\tiny$\pm 0.60$} & \textbf{76.6}{\tiny$\pm 0.78$} \\

    \midrule
    \vspace{-0.8em} \\ 
    \midrule
    \multicolumn{13}{c}{\textbf{ACE05-E Test Set}} \\
    \midrule
    \multirow{2}{*}{Model} & \multicolumn{2}{c|}{5\%} & \multicolumn{2}{c|}{10\%} & \multicolumn{2}{c|}{20\%} & \multicolumn{2}{c|}{30\%} & \multicolumn{2}{c|}{50\%} & \multicolumn{2}{c}{100\%} \\
    \cmidrule{2-13}
    & Arg-I     & Arg-C     & Arg-I     & Arg-C     & Arg-I     & Arg-C     & Arg-I     & Arg-C     & Arg-I     & Arg-C     & Arg-I     & Arg-C \\
    \midrule
    DyGIE++~\cite{Wadden19dygiepp}
    & 39.2{\tiny$\pm 4.20$} & 29.3{\tiny$\pm 2.63$} & 50.5{\tiny$\pm 1.44$} & 42.2{\tiny$\pm 0.85$} & 57.7{\tiny$\pm 1.11$} & 49.5{\tiny$\pm 0.75$} & 59.9{\tiny$\pm 0.97$} & 53.2{\tiny$\pm 1.38$} & 61.0{\tiny$\pm 2.62$} & 54.4{\tiny$\pm 1.10$} & 63.6{\tiny$\pm 1.74$} & 57.4{\tiny$\pm 1.87$} \\
    OneIE~\cite{Lin20oneie}
     & 41.3{\tiny$\pm 1.97$} & 34.6{\tiny$\pm 1.88$} & 55.4{\tiny$\pm 2.29$} & 50.0{\tiny$\pm 1.51$} & 64.6{\tiny$\pm 2.54$} & 59.6{\tiny$\pm 1.12$} & 67.8{\tiny$\pm 1.50$} & 63.0{\tiny$\pm 1.43$} & 72.0{\tiny$\pm 0.43$} & 68.3{\tiny$\pm 0.92$} & 73.7{\tiny$\pm 0.87$} & 70.7{\tiny$\pm 0.38$} \\
    Query and Extract~\cite{wang-etal-2022-query}
    & 36.8{\tiny$\pm 3.44$} & 11.0{\tiny$\pm 0.50$} & 33.1{\tiny$\pm 5.45$} & 20.9{\tiny$\pm 2.83$} & 45.6{\tiny$\pm 0.93$} & 34.3{\tiny$\pm 1.30$} & 51.1{\tiny$\pm 3.78$} & 44.3{\tiny$\pm 4.01$} & 56.1{\tiny$\pm 4.87$} & 49.6{\tiny$\pm 5.15$} & 62.4{\tiny$\pm 2.10$} & 59.1{\tiny$\pm 1.88$} \\
    AMR-IE~\cite{DBLP:conf/naacl/ZhangJ21}
    & 43.2{\tiny$\pm 1.54$} & 36.8{\tiny$\pm 0.07$} & 53.3{\tiny$\pm 1.49$} & 48.5{\tiny$\pm 0.99$} & 63.2{\tiny$\pm 0.60$} & 58.3{\tiny$\pm 0.93$} & 67.2{\tiny$\pm 1.00$} & 62.6{\tiny$\pm 1.16$} & 69.5{\tiny$\pm 1.27$} & 66.1{\tiny$\pm 0.92$} & 73.6{\tiny$\pm 0.40$} & 70.3{\tiny$\pm 0.13$} \\
    PAIE~\cite{ma-etal-2022-prompt}
    & 52.2{\tiny$\pm 0.83$} & 46.3{\tiny$\pm 0.75$} & 62.0{\tiny$\pm 0.96$} & 56.3{\tiny$\pm 0.46$} & 67.8{\tiny$\pm 0.33$} & 62.8{\tiny$\pm 0.69$} & 71.3{\tiny$\pm 0.54$} & 65.8{\tiny$\pm 0.98$} & 72.8{\tiny$\pm 2.34$} & 69.1{\tiny$\pm 2.20$} & 75.0{\tiny$\pm 0.51$} & 72.1{\tiny$\pm 0.69$} \\
    \degree{}~\cite{Hsu22degree}
    & 47.7{\tiny$\pm 0.09$} & 41.7{\tiny$\pm 0.83$} & 63.0{\tiny$\pm 1.45$} & 57.7{\tiny$\pm 1.72$} & 64.2{\tiny$\pm 0.57$} & 58.9{\tiny$\pm 1.00$} & 70.3{\tiny$\pm 1.16$} & 65.8{\tiny$\pm 1.50$} & 71.4{\tiny$\pm 0.26$} & 68.2{\tiny$\pm 0.25$} & 75.6{\tiny$\pm 0.79$} & 73.0{\tiny$\pm 0.53$} \\
    \midrule
    \modelbart{} 
     & \ul{59.8}{\tiny$\pm 2.00$} & \ul{52.4}{\tiny$\pm 1.53$} & \ul{66.0}{\tiny$\pm 1.82$} & \ul{61.0}{\tiny$\pm 1.58$} & \ul{70.2}{\tiny$\pm 0.84$} & \textbf{66.4}{\tiny$\pm 1.04$} & \textbf{73.3}{\tiny$\pm 0.45$} & \textbf{69.7}{\tiny$\pm 0.41$} & \ul{74.4}{\tiny$\pm 1.21$} & \ul{71.1}{\tiny$\pm 1.17$} & \ul{76.0}{\tiny$\pm 0.85$} & \ul{73.4}{\tiny$\pm 0.58$}\\
    \modelroberta{} 
    & \textbf{61.0}{\tiny$\pm 0.98$} & \textbf{53.4}{\tiny$\pm 0.21$} & \textbf{67.8}{\tiny$\pm 1.13$} & \textbf{61.7}{\tiny$\pm 0.79$} & \textbf{70.5}{\tiny$\pm 1.28$} & \textbf{66.4}{\tiny$\pm 0.86$} & \ul{73.1}{\tiny$\pm 0.43$} & \ul{69.5}{\tiny$\pm 0.67$} & \textbf{74.6}{\tiny$\pm 1.03$} & \textbf{71.9}{\tiny$\pm 0.89$} & \textbf{76.7}{\tiny$\pm 0.75$} & \textbf{74.2}{\tiny$\pm 0.28$}\\
    
    \midrule
    \vspace{-0.8em} \\ 
    \midrule
    \multicolumn{13}{c}{\textbf{ERE-EN Development Set}} \\
    \midrule
    \multirow{2}{*}{Model} & \multicolumn{2}{c|}{5\%} & \multicolumn{2}{c|}{10\%} & \multicolumn{2}{c|}{20\%} & \multicolumn{2}{c|}{30\%} & \multicolumn{2}{c|}{50\%} & \multicolumn{2}{c}{100\%} \\
    \cmidrule{2-13}
    & Arg-I     & Arg-C     & Arg-I     & Arg-C     & Arg-I     & Arg-C     & Arg-I     & Arg-C     & Arg-I     & Arg-C     & Arg-I     & Arg-C \\
    \midrule
    DyGIE++~\cite{Wadden19dygiepp}
     & 51.8{\tiny$\pm 2.16$} & 42.2{\tiny$\pm 0.68$} & 52.9{\tiny$\pm 3.19$} & 45.4{\tiny$\pm 2.65$} & 56.8{\tiny$\pm 1.84$} & 49.0{\tiny$\pm 0.58$} & 57.3{\tiny$\pm 0.67$} & 50.1{\tiny$\pm 0.96$} & 58.8{\tiny$\pm 0.56$} & 51.5{\tiny$\pm 1.47$} & 63.8{\tiny$\pm 2.20$} & 56.8{\tiny$\pm 1.93$} \\
    OneIE~\cite{Lin20oneie}
    & 56.8{\tiny$\pm 3.14$} & 51.4{\tiny$\pm 2.58$} & 65.6{\tiny$\pm 0.42$} & 59.5{\tiny$\pm 0.71$} & 68.6{\tiny$\pm 0.55$} & 62.0{\tiny$\pm 0.56$} & 70.8{\tiny$\pm 0.75$} & 65.5{\tiny$\pm 0.51$} & 73.6{\tiny$\pm 0.56$} & 68.5{\tiny$\pm 0.67$} & 75.5{\tiny$\pm 0.26$} & 71.2{\tiny$\pm 0.13$} \\
    Query and Extract~\cite{wang-etal-2022-query}
    & 34.8{\tiny$\pm 6.37$} & 22.0{\tiny$\pm 5.30$} & 45.9{\tiny$\pm 1.59$} & 37.3{\tiny$\pm 2.03$} & 49.0{\tiny$\pm 4.31$} & 41.2{\tiny$\pm 3.38$} & 56.1{\tiny$\pm 1.32$} & 49.4{\tiny$\pm 1.19$} & 63.5{\tiny$\pm 1.81$} & 57.0{\tiny$\pm 1.34$} & 70.4{\tiny$\pm 2.17$} & 65.0{\tiny$\pm 2.16$} \\
    AMR-IE~\cite{DBLP:conf/naacl/ZhangJ21}
    & 48.4{\tiny$\pm 1.48$} & 44.8{\tiny$\pm 0.86$} & 61.2{\tiny$\pm 0.97$} & 55.2{\tiny$\pm 1.06$} & 63.0{\tiny$\pm 1.37$} & 56.9{\tiny$\pm 1.12$} & 70.5{\tiny$\pm 0.03$} & 65.2{\tiny$\pm 0.63$} & 73.0{\tiny$\pm 0.79$} & 67.6{\tiny$\pm 0.39$} & 75.3{\tiny$\pm 1.30$} & 70.1{\tiny$\pm 1.45$} \\
    \degree{}~\cite{Hsu22degree}
     & 64.2{\tiny$\pm 0.33$} & 57.2{\tiny$\pm 0.21$} & 69.7{\tiny$\pm 0.36$} & 62.5{\tiny$\pm 0.89$} & 69.2{\tiny$\pm 0.42$} & 63.9{\tiny$\pm 0.55$} & 73.4{\tiny$\pm 0.35$} & 67.1{\tiny$\pm 0.11$} & 75.4{\tiny$\pm 0.52$} & 70.2{\tiny$\pm 0.48$} & 77.4{\tiny$\pm 0.32$} & 73.3{\tiny$\pm 0.52$} \\
    \midrule
    \modelbart{} 
    & \ul{69.2}{\tiny$\pm 1.64$} & \ul{62.4}{\tiny$\pm 1.54$} & \textbf{72.8}{\tiny$\pm 1.12$} & \textbf{66.8}{\tiny$\pm 1.03$} & \ul{71.5}{\tiny$\pm 0.51$} & \ul{66.0}{\tiny$\pm 0.95$} & \textbf{74.9}{\tiny$\pm 0.65$} & \ul{68.8}{\tiny$\pm 0.17$} & \textbf{76.7}{\tiny$\pm 0.33$} & \textbf{70.8}{\tiny$\pm 0.55$} & \textbf{78.1}{\tiny$\pm 0.69$} & \ul{73.6}{\tiny$\pm 1.10$} \\
    \modelroberta{} 
    & \textbf{69.9}{\tiny$\pm 0.97$} & \textbf{63.1}{\tiny$\pm 1.24$} & \ul{72.7}{\tiny$\pm 0.81$} & \ul{66.7}{\tiny$\pm 0.56$} & \textbf{71.7}{\tiny$\pm 0.33$} & \textbf{66.6}{\tiny$\pm 0.98$} & \ul{74.6}{\tiny$\pm 0.52$} & \textbf{69.7}{\tiny$\pm 0.68$} & \ul{75.7}{\tiny$\pm 0.74$} & \ul{70.6}{\tiny$\pm 0.67$} & \ul{77.9}{\tiny$\pm 0.28$} & \textbf{73.8}{\tiny$\pm 0.34$} \\

    \midrule
    \vspace{-0.8em} \\ 
    \midrule
    \multicolumn{13}{c}{\textbf{ERE-EN Test Set}} \\
    \midrule
    \multirow{2}{*}{Model} & \multicolumn{2}{c|}{5\%} & \multicolumn{2}{c|}{10\%} & \multicolumn{2}{c|}{20\%} & \multicolumn{2}{c|}{30\%} & \multicolumn{2}{c|}{50\%} & \multicolumn{2}{c}{100\%} \\
    \cmidrule{2-13}
    & Arg-I     & Arg-C     & Arg-I     & Arg-C     & Arg-I     & Arg-C     & Arg-I     & Arg-C     & Arg-I     & Arg-C     & Arg-I     & Arg-C \\
    \midrule
    DyGIE++~\cite{Wadden19dygiepp}
     & 53.3{\tiny$\pm 1.95$} & 40.0{\tiny$\pm 1.93$} & 52.9{\tiny$\pm 2.59$} & 44.6{\tiny$\pm 2.70$} & 55.9{\tiny$\pm 1.74$} & 49.5{\tiny$\pm 1.16$} & 59.1{\tiny$\pm 0.64$} & 52.0{\tiny$\pm 1.35$} & 60.5{\tiny$\pm 0.92$} & 53.7{\tiny$\pm 0.38$} & 63.4{\tiny$\pm 0.80$} & 56.0{\tiny$\pm 0.78$} \\
    OneIE~\cite{Lin20oneie}
    & 55.5{\tiny$\pm 3.47$} & 49.5{\tiny$\pm 2.24$} & 62.1{\tiny$\pm 1.53$} & 56.1{\tiny$\pm 1.62$} & 67.9{\tiny$\pm 1.83$} & 62.3{\tiny$\pm 1.62$} & 71.9{\tiny$\pm 0.36$} & 66.1{\tiny$\pm 0.73$} & 72.3{\tiny$\pm 0.49$} & 67.7{\tiny$\pm 0.43$} & 75.2{\tiny$\pm 1.14$} & 70.1{\tiny$\pm 1.96$} \\
    Query and Extract~\cite{wang-etal-2022-query}
     & 35.1{\tiny$\pm 7.25$} & 19.7{\tiny$\pm 5.12$} & 46.7{\tiny$\pm 2.66$} & 34.0{\tiny$\pm 4.06$} & 52.1{\tiny$\pm 4.69$} & 42.4{\tiny$\pm 5.07$} & 57.7{\tiny$\pm 0.09$} & 50.1{\tiny$\pm 0.86$} & 64.5{\tiny$\pm 2.78$} & 57.7{\tiny$\pm 2.80$} & 70.4{\tiny$\pm 1.78$} & 64.3{\tiny$\pm 2.26$} \\
    AMR-IE~\cite{DBLP:conf/naacl/ZhangJ21}
    & 47.8{\tiny$\pm 0.65$} & 44.1{\tiny$\pm 0.46$} & 59.1{\tiny$\pm 0.96$} & 53.7{\tiny$\pm 0.58$} & 65.8{\tiny$\pm 1.68$} & 60.4{\tiny$\pm 1.22$} & 71.4{\tiny$\pm 1.31$} & 65.7{\tiny$\pm 1.45$} & 73.9{\tiny$\pm 0.44$} & 68.8{\tiny$\pm 0.29$} & 76.5{\tiny$\pm 1.20$} & 71.5{\tiny$\pm 1.34$} \\
    \degree{}~\cite{Hsu22degree}
      & 66.4{\tiny$\pm 0.14$} & 57.5{\tiny$\pm 0.36$} & 71.2{\tiny$\pm 1.26$} & 63.9{\tiny$\pm 1.38$} & 72.3{\tiny$\pm 0.69$} & 67.4{\tiny$\pm 0.56$} & 74.1{\tiny$\pm 1.16$} & 69.1{\tiny$\pm 1.44$} & \textbf{77.4}{\tiny$\pm 0.61$} & \textbf{73.3}{\tiny$\pm 0.74$} & 78.2{\tiny$\pm 0.69$} & 74.9{\tiny$\pm 1.10$} \\
    \midrule
    \modelbart{} 
    & \ul{71.3}{\tiny$\pm 0.40$} & \ul{62.9}{\tiny$\pm 0.53$} & \ul{73.7}{\tiny$\pm 0.73$} & \ul{66.7}{\tiny$\pm 0.45$} & \ul{73.1}{\tiny$\pm 0.57$} & \textbf{68.5}{\tiny$\pm 0.44$} & \textbf{75.7}{\tiny$\pm 0.83$} & \textbf{71.3}{\tiny$\pm 0.70$} & \ul{77.1}{\tiny$\pm 0.30$} & \ul{72.5}{\tiny$\pm 1.07$} & \textbf{78.8}{\tiny$\pm 0.62$} & \textbf{75.4}{\tiny$\pm 0.59$} \\
    \modelroberta{} 
    & \textbf{71.4}{\tiny$\pm 1.13$} & \textbf{63.2}{\tiny$\pm 0.57$} & \textbf{73.8}{\tiny$\pm 0.57$} & \textbf{67.7}{\tiny$\pm 0.66$} & \textbf{73.6}{\tiny$\pm 0.64$} & \ul{68.4}{\tiny$\pm 0.40$} & \ul{75.4}{\tiny$\pm 0.36$} & \ul{70.5}{\tiny$\pm 0.17$} & 77.0{\tiny$\pm 0.73$} & \ul{72.5}{\tiny$\pm 0.68$} & \ul{78.4}{\tiny$\pm 0.80$} & \ul{75.0}{\tiny$\pm 0.77$} \\

    \bottomrule
\end{tabular}
}
\caption{Argument Identification and classification F1-scores (\%) under different data proportion settings for ACE05-E and ERE-EN datasets. The highest scores are in bold and the second-best scores are underlined. The reported numbers are the average of the results from three random runs. The standard deviation (\%) of three runs are also reported in the table. }
\label{table:main_result_detail}
\end{table*}